\documentclass[10pt,twocolumn,letterpaper]{article}

\usepackage{cvpr}              % To produce the CAMERA-READY version
\usepackage{graphicx, nicefrac}
\usepackage{amsmath}
\usepackage{amssymb}
\usepackage{booktabs}
\usepackage{multirow}
\usepackage{mathrsfs}
\usepackage{subcaption}
\usepackage{pifont}

\newcommand{\xmark}{\ding{55}}%

\usepackage[pagebackref,breaklinks,colorlinks]{hyperref}

\usepackage[capitalize]{cleveref}
\crefname{section}{Sec.}{Secs.}
\Crefname{section}{Section}{Sections}
\Crefname{table}{Table}{Tables}
\crefname{table}{Tab.}{Tabs.}

\def\cvprPaperID{69} % *** Enter the CVPR Paper ID here
\def\confName{CVPR}
\def\confYear{2023}

\begin{document}

%%%%%%%%% TITLE - PLEASE UPDATE
%\title{TTT-UCDR: Test-time Training for Universal Cross-Domain Retrieval}
\title{Test-time Training for Data-efficient UCDR}

% For a paper whose authors are all at the same institution,
% omit the following lines up until the closing ``}''.
% Additional authors and addresses can be added with ``\and'',
% just like the second author.
% To save space, use either the email address or home page, not both
\author{Soumava Paul\textsuperscript{1}, Titir Dutta\textsuperscript{2},  Aheli Saha\textsuperscript{1}, Abhishek Samanta\textsuperscript{1}, Soma Biswas\textsuperscript{2}\\
{\textsuperscript{1}Universit{\"a}t des Saarlandes, } { \textsuperscript{2}Indian Institute of Science, Bangalore} \\
{\tt\small \textsuperscript{1}\{sopa00001, ahsa00002, absa00002\}@stud.uni-saarland.de, \textsuperscript{2}\{titird,somabiswas\}@iisc.ac.in}
}
\maketitle

%%%%%%%%% ABSTRACT
\begin{abstract}
   Image retrieval under generalized test scenarios has gained significant momentum in literature, and the recently proposed protocol of Universal Cross-domain Retrieval is a pioneer in this direction. A common practice in any such generalized classification or retrieval algorithm is to exploit samples from many domains during training to learn a domain-invariant representation of data. Such criterion is often restrictive, and thus in this work, for the first time, we explore the generalized retrieval problem in a data-efficient manner. Specifically, we aim to generalize any pre-trained cross-domain retrieval network towards any unknown query domain/category, by means of adapting the model on the test data leveraging self-supervised learning techniques. Toward that goal, we explored different self-supervised loss functions~(for example, RotNet, JigSaw, Barlow Twins, etc.) and analyze their effectiveness for the same. Extensive experiments demonstrate the proposed approach is simple, easy to implement, and effective in handling data-efficient UCDR.
   %We attempt to make any retrieval model trained on a small cross-domain dataset (containing just two training domains) more generalizable towards any unknown query domain or category by quickly adapting it to the test data during inference. This form of test-time training or adaptation of the retrieval model is explored by means of a number of self-supervision-based loss functions, for example, Rotnet, Jigsaw-puzzle, Barlow twins, etc., in this work. Extensive experiments on multiple large-scale datasets demonstrate the effectiveness of the proposed approach.
\end{abstract}

%%%%%%%%% BODY TEXT
\section{Introduction}
\label{sec:intro}

Cross-domain data retrieval has emerged as an important and quite relevant research topic in today's world because of the plethora of information being uploaded and shared through internet in multiple forms or modalities (text, video, image etc.) and categories. 
A number of research has been going on to address this problem, but mostly with a pre-defined domain of retrieval, such as text-based image retrieval~\cite{textImageretrieval}, sketch-based image retrieval~\cite{liu2017deep}, audio-based object retrieval~\cite{arandjelovic2018objects} etc.
Even though all of these retrieval scenarios have important individual applications in real life, in recent times, we have observed these domain boundaries to be blurrier than ever. 
For example, Google has a search platform which can process user input or queries in the forms of keywords~(text-based retrieval), voice commands ~(audio-based retrieval), and image, simultaneously.
Maintaining domain-specific retrieval networks for each of these could lead to huge training and maintenance cost.
%to process each of these input domains for such a multi-domain information retrieval system is a solution, which can result in high costs and a huge volume of training data requirement to maintain the quality of retrieval.
%This indicates the need for a generalized model that can seamlessly handle such user queries from multiple or unknown domains.
Thus we see an initiative to address such problems in terms of domain-agnostic retrieval models, which is formally introduced as Universal Cross-domain Retrieval~(UCDR) in~\cite{paul2021universal}.
UCDR attempts to combine the Domain Generalization efforts for classification~\cite{Cumix} with the traditional Zero-shot Sketch-based Image Retrieval~(ZS-SBIR)~\cite{yelamarthi2018zero}\cite{shen2018zero}. 
The query in this case, can belong to an unseen category, as well as can be from a unseen domain; and thus the retrieval scenario becomes relatable to real-world and much more challenging compared to the stand-alone DG or ZS-SBIR problem.

However, to address such generalization across domains, both DG and UCDR methods combine training data collected over multiple domains~(of the same set of categories) to learn semantically-meaningful domain-independent representation, which can translate directly to any unknown domain during retrieval.
Clearly, this imposes restrictions on the models to be built with multi-domain, multi-category training samples of huge quantities.
Additionally, any domain-specific efficient retrieval model~(such as DSH~\cite{DSH} for SBIR), with a comparatively lower training data~(only sketch and image domains) requirement, becomes irrelevant in this case, since those would be heavily biased towards sketch and image domains only.
This is extremely restrictive in real-world, since collecting annotated data from multiple domains involves significant manual labour.
Thus, we feel that it is time to take a step back and analyze the possibility to achieve such generalization in a data-efficient manner.

In this work, we aim to explore a data-efficient methodology to address the universal cross-domain retrieval problem.
Specifically, we attempt to explore the possibility of re-using any cross-domain retrieval network, trained on a comparatively smaller set~(compared to DG or UCDR trainings) and adapt them to address retrieval under UCDR protocol.
Towards this goal, we choose the Semantic-neighbourhood and Mixture-prediction network~(SnMpNet), originally proposed to address UCDR, as the baseline model.
We first explore the effect on its performance when trained with only two domains, eg. sketch and image~(instead of five, as shown in~\cite{paul2021universal}).
Next, we attempt to adapt the model, trained in a data-scarce setting, by leveraging the test-time samples from an unknown category and unknown domain, to improve its performance.
This approach is based on our hypothesis that any information extracted from a query sample of an unknown domain and/or category may help an already trained model adapt to the underlying distribution shift quickly and thus result in a better retrieval list.
Such adaptation is inspired from the test-time training~(TTT)~\cite{sun2020test} strategy, originally proposed for classification, and later applied for ZS-SBIR in~\cite{sain2022sketch3t}.
%of such retrieval model.
TTT proposes to update an already trained model with the information extracted from test samples during test-time, by minimizing a self-supervised loss function.
In this work, we explore three different self-supervised losses for the same, namely - (1) RotNet loss~\cite{gidaris2018unsupervised}, (2) Jigsaw puzzle loss~\cite{noroozi2016unsupervised}, and (3) Barlow Twins~\cite{zbontar2021barlow} loss.
It can be noted that, unlike the original TTT~\cite{sun2020test}, our proposed methodology~(details in Section~\ref{sec_proposed}) does not require the inclusion of these self-supervised losses during original training of the baseline model.
Thus our test-time adaptation process is much simpler, and easy to implement.
Additionally, this adaptation can be seamlessly incorporated with any existing retrieval algorithm, without modifying their original training or architecture.

Thus, we summarize the contributions of this work as follows:
\begin{enumerate}
    \item In this work, we attempt to explore the UCDR problem in a data-efficient manner, instead of training on a large-scale multi-domain dataset. %SnMpNet~\cite{paul2021universal} is treated as the base model.
    \item We explore a number of self-supervision based learning techniques to adapt the pre-trained base model towards unseen query data.
%    \item We also explore the effectiveness of the test-time training strategy in UCDR setting.
    \item Extensive experiments and analysis on the large-scale DomainNet dataset are performed to demonstrate the effectiveness of the proposed training and adaptation strategy.
\end{enumerate}
Next, we briefly discuss the relevant recent work in this direction in Section~\ref{sec_reference}.
The rest of the paper is organized in the following fashion: we discuss the proposed test-time training approach using self-supervised losses in Section~\ref{sec_proposed}, followed by our findings and analysis in Sections~\ref{sec_results} and~\ref{sec_analysis}. 
Finally, we conclude this paper with a summary in Section~\ref{sec_conclusion}.

%We further demonstrate the benefits of test-time training in a particularly challenging cross-dataset setting. Extensive experiments on the large-scale Sketchy~\cite{Sketchy} and DomainNet~\cite{Domainnet} datasets and comparisons with \emph{SnMpNet} show the effectiveness of our methods.

\begin{figure*}
    \includegraphics[width=1.0\linewidth]{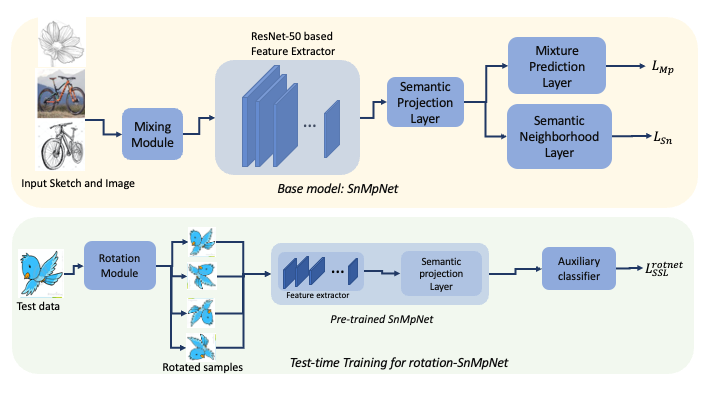}
    \caption{This figure depicts test-time training strategy for rotation-SnMpNet in data-efficient UCDR.}
    \label{fig_network}
\end{figure*}

%------------------------------------------------------------------------
\section{Related Work}
\label{sec_reference}
Here, we discuss some of the seminal works in image retrieval, test-time training, and self-supervised representation learning to elaborate on the background of this paper.
%\textbf{Sketch-based Image Retrieval~(SBIR): }
%The objective of early SBIR methods, e.g. HOG~\cite{HOG}, LKS~\cite{LKS} was to learn hand-crafted representations of sketch and images so that the domain-gap is minimized.
%This approach has later been replaced by the deep learning methodologies, such as Siamese network~\cite{Siamese}, GoogleNet~\cite{Sketchy}.
%Recent hashing-based methods, DSH~\cite{DSH}, GDH~\cite{GDH} introduce the  domain-restricted two-branched architecture for SBIR, which learns a shared latent-space representation and perform retrieval in that space.

\textbf{Zero-shot Sketch-based Image Retrieval~(ZS-SBIR): }
It addresses the problem of sketch-based image retrieval when the query sketch belongs to a category that was not seen by the model during training.
The problem was first introduced by \cite{shen2018zero}\cite{yelamarthi2018zero}, as a category-wise generalization extension of traditional sketch-based image retrieval~(SBIR)~\cite{liu2017deep}\cite{zhang2018generative}.
Later, \cite{dey2019doodle}\cite{dutta2019semantically}\cite{dutta2020adaptive}\cite{dutta2020styleguide} have reported significant improvements in this direction.
The general approach for this problem followed in these papers is to learn a latest-space or shared-space representation of sketch and images, by means of a semantic supervision~(generally the word2vec or GloVe-representation of the training category names), so that the sketch and images from same categories can be placed close to each other in this learned latent-space.
In contrast to this popular approach, \cite{liu2019semantic} proposed a single-branch architecture for both domains with a domain-indicator function to reduce the number of trainable parameters.
This architecture has been adopted very closely in our base model SnMpNet~\cite{paul2021universal} for UCDR.
%This problem refers to the retrieval of unseen classes using sketches as queries, to bridge the gap between the availability of real images and sketches, and was first introduced in \cite{shen2018zero}\cite{yelamarthi2018zero}. This is an extension of Sketch-based Image Retrieval (SBIR) \cite{liu2017deep}\cite{zhang2018generative}, where all classes are seen during training, and has been widely addressed by various algorithms \cite{dey2019doodle}\cite{dutta2019semantically}\cite{dutta2020adaptive}\cite{dutta2020styleguide}. Opposed to the two-parallel branch architecture common in literature, \cite{liu2019semantic} introduced a network with a single branch and domain-indicator function to incorporate domain knowledge.

\textbf{Universal Cross-Domain Retrieval~(UCDR): }
UCDR-protocol \cite{paul2021universal} further extends ZS-SBIR towards domain-wise generalization during retrieval.
Thus, it proposed to move beyond sketch-query, successfully modeling any real-life retrieval scenario.
The proposed model SnMpNet for UCDR learns a domain-invariant and semantically meaningful representation of data for retrieval using a single-branch architecture as in~\cite{liu2019semantic}.
The details of this model have been discussed in Section~\ref{sec_base_model}.
%UCDR \cite{paul2021universal} further extends ZS-SBIR to encompass ZS-SBIR as well as Domain Generalization (DG), thus moving beyond the restrictions of the sketch domain. It tackles an even more realistic framework, where the query sample can belong to an unseen class as well as an unseen domain. It aims to learn a domain-independent semantic representation for each class, and to this end proposes two losses - Semantic Neighbourhood Loss, accounting for unobserved classes, and Mixture Prediction Loss, accounting for hidden domains. These, alongside a single-branch model, constitute \emph{SnMpNet}.

\textbf{Test-Time Training~(TTT): }
This was first introduced by \cite{sun2020test}, where the test data is treated as an unlabelled dataset, and the model weights are updated via a self-supervised learning approach. 
It aims at improving model generalization by increasing  robustness against distribution shifts, owing to real-life situations where the train and test data often belong to different distributions. 
The rotation prediction task proposed by \cite{gidaris2018unsupervised} is used as the learning objective for this self-supervision task. 
Recently,~\cite{ttt++} performed a detailed analysis of such test-time training strategies under significant distribution shift, and proposed a test-time feature-alignment and moment-matching strategy to address the same.
Tent~\cite{wang2020tent} adapts the model parameters to distribution shifts in test-time by minimizing the entropy of model predictions. 
However, the entropy minimization loss can only be used in the classification setting since it belongs to a class of probabilistic losses, which is not present in the retrieval setting of SnMpNet. 
\cite{bartler2022mt3} combines meta-learning, self-supervision, and test-time adaptation to address corrupted image classification benchmark on the CIFAR-10 dataset.
A contrastive self-supervision learning technique is combined with pseudo-labeling in~\cite{chen2022contrastive}. 
\cite{wang2022continual} proposes a continual learning technique for test-time adaptation.
Sketch3T~\cite{sain2022sketch3t} utilizes a self-supervised task of sketch-raster to sketch vector translations. This helps in adapting the model to the unique style of new sketches in test-time, as well as new categories. 
This is the first work to apply test-time training to the ZS-SBIR task. 
However, it deals solely with the sketch domain. 
In the UCDR setting, the query sample can belong to any unseen domain, and hence it's not possible to use such a specialized self-supervised loss for test-time adaptation. 

\textbf{Self-Supervised Learning: }
Recent self-supervised algorithms share a common methodology of learning semantic information about data, and are independent of variations in style, orientation, and distortions. 
Few notable works in this direction are RotNet \cite{gidaris2018unsupervised}, Jigsaw Puzzles \cite{noroozi2016unsupervised}, Barlow Twins \cite{zbontar2021barlow}, \cite{grill2020bootstrap}, \cite{chen2020simple} etc. 
RotNet~\cite{gidaris2018unsupervised} argues that a model capable of predicting the rotational angles applied to an image necessarily has contextual and class awareness, and therefore the rotation prediction task can be employed for self-supervised representation learning.
Jigsaw Puzzles \cite{noroozi2016unsupervised} formulates the self-supervision learning objective as a jigsaw puzzle-solving task to gain visuospatial understanding.
Barlow Twins \cite{zbontar2021barlow} proposes a cross-correlation matrix between representations of distorted versions of the same batch as an objective function and aims to drive it toward an identity matrix.
BYOL~\cite{grill2020bootstrap} proposes two neural networks, online and target to interact and learn from each other while processing augmented versions of the same sample.
SimCLR~\cite{chen2020simple} proposes a contrastive self-supervision learning technique without requiring a memory bank.
It is to be noted that the overall objective of self-supervision learning techniques resonates well with the goal of UCDR, where we want to learn domain-independent representations of a class.
Thus it makes for a logical choice to achieve generalization from unknown test data.

%This resonates well with the goal of UCDR, where we want to learn domain-independent representations of a class. 
%We, therefore, incorporate the following three well-known self-supervised algorithms for test-time training: 
%1) RotNet \cite{gidaris2018unsupervised} - This work argues that a model capable of predicting the rotational angles applied to an image necessarily has contextual and class awareness, and therefore the rotation prediction task(0°, 90°, 180°, 270°) can be employed for self-supervised representation learning.
%2) Jigsaw Puzzles \cite{noroozi2016unsupervised} - This work formulates the self-supervision learning objective as a jigsaw puzzle-solving task to gain visuospatial understanding. Nine tiles are extracted from the images and shuffled into one of thirty identifiable patterns (out of 9! possible permutations), which is then predicted by the model.
%3) Barlow Twins \cite{zbontar2021barlow} - This work proposes a cross-correlation matrix between representations of distorted versions of the same batch as an objective function and aims to drive it toward an identity matrix.

%------------------------------------------------------------------------
\section{Proposed Method}
\label{sec_proposed}
We begin the discussion of the proposed method with a short description of the base model SnMpNet~\cite{paul2021universal}.

\subsection{Base Model - SnMpNet: }
\label{sec_base_model}
The Semantic Neighbourhood and Mixture Prediction Network~(SnMpNet) has a deep neural architecture, with SE-ResNet50 as its backbone feature extractor. 
%\color{red} embedding dimension \color{black}
The network implies a semantic feature~(through a linear projection layer) of 300-d, on top of the SE-ResNet~(2048-d).
SnMpNet aims to learn this semantic feature to be domain-invariant, as well as meaningful, so that domain-generalization and category-wise generalization can be achieved simultaneously.
Following CuMix~\cite{Cumix}, SnMpNet also treats the input data in a mixed format, where the mixing may be performed inter or intra-domain.
Thus, for given samples $\mathbf{x}_i$, $\mathbf{x_j}$ and $\mathbf{x}_k$ from training set $\mathcal{D}_{tr} = \{(\mathbf{x}_m, y_m)\}_{m=1}^{N}$ of $N$-samples, the input to the network is computed as, $\mathbf{x} = \alpha \mathbf{x}_i + (1-\alpha)[\lambda \mathbf{x}_j + (1-\lambda) \mathbf{x}_k]$.
Here, $\alpha \sim Beta(\beta,\beta)$ and $\lambda \sim Bernoulli(\gamma, \gamma)$, with $\beta$ and $\gamma$ being hyper-parameters.
To obtain a domain-invariant representation of such mixed samples, the mixture-prediction loss ~$\mathcal{L}_{Mp}$ is introduced, which only predicts the correct ratio of the component categories present in $\mathbf{x}$, and \emph{forgets} about their domain.
%\begin{align}
%    \mathcal{L}_{Mp} &= 
%\end{align}
%By predicting only the component categories in this mixed sample, essentially, the model tries to \emph{forget} the domain-specific information present in it, thus producing a domain-invariant representation of data.
Additionally, SnMpNet also minimizes a semantic neighbourhood loss $\mathcal{L}_{Sn}$, which is essentially a cross-entropy loss computed between the latent-space representation of $\mathbf{x}$ and its components' semantic ground-truth data~(eg., word2vec representations of their category-name).
Combining both these losses, the network learns meaningful semantic-representation of data, which are domain-agnostic.
%This is a generic approach followed by many ZS-SBIR algorithms to obtain a semantically meaningful representation of unseen category data, which can bring the relevant sketch and images close to each other in the representation space.
Final retrieval is performed in the learned semantic-space on the basis of the Euclidean distance between the query sample and search-set instances.

In our work, we retain the architecture and training methodology of SnMpNet unmodified, which provisions for this base model to be replaced by any other cross-domain retrieval algorithm with a shared-space representation learning technique.
Next, we discuss the test-time training proposed on top of such a retrieval algorithm to enhance its performance for UCDR. We begin this discussion by providing a brief insight into the self-supervision techniques explored for the same.

\subsection{Self-supervision Techniques for Unknown Data}
Self-supervised learning~(SSL) has become a popular choice when learning from unlabeled or unstructured data.
Here we specifically explore the following three SSL-loss components in this regard.
We choose RotNet~\cite{gidaris2018unsupervised} and Jigsaw\cite{noroozi2016unsupervised} because of their simplicity, as well as Barlow twins~\cite{zbontar2021barlow} for its effectiveness. 
%\textcolor{red}{in the low-data condition}.
We first discuss the details on the loss functions and then describe the adaptation process followed using each of these losses in this work.
% Here, we first describe the losses.

\textbf{RotNet Loss: }
RotNet~\cite{gidaris2018unsupervised} has been a very popular choice for introducing self-supervision in a feature learning network, due to its simplicity and effectiveness.
It uses four rotations of an input image, \textcolor{black}{in angles of $\{0^{\circ}, 90^{\circ}, 180^{\circ}, 270^{\circ}\}$}, and learns to predict the rotation-index of any such sample from its corresponding feature representations.
Thus, the RotNet-loss is computed as,
\begin{align}
    \mathcal{L}_{SSL}^{rotnet} &= \frac{1}{n} \sum_{i=1}^{n}{\mathcal{L}_{CE}(r_i, h_i)}
\end{align}
where $\mathcal{L}_{CE}$ is the cross-entropy loss computed between the true rotation-index $r_i$ and the predicted index $h_i$.
The loss is averaged over the total $n$-number of samples present in the \color{black}test set. \color{black}
This loss is minimized without the direct class supervision of the samples, but indirectly learns the semantic content of the data through the rotation prediction. \\

\textbf{Jigsaw Puzzles Loss: }
Similar to RotNet, jigsaw puzzle~\cite{noroozi2016unsupervised} is another very effective self-supervision component, \textcolor{black}{which has been used in transfer learning~\cite{noroozi2016unsupervised}, domain-generalization}~\cite{eisnet} etc. %\textcolor{red}{many applications of such as domain-generalization}~\cite{eisnet}}.
Here, any input image is broken down into a number of patches, based on the fixed grid size. 
For example, following the authors' approach in~\cite{noroozi2016unsupervised}, we resize and then break down an image into $9$-patches using a $3\times3$ grid. 
These $9$-patches are then shuffled and a total of $31$ different combinations (out of a possible $9!$) of \emph{jigsaw} images are created through permutations.
Now, the network is trained to predict the permutation index of such jigsaw images, forming the following jigsaw-puzzle loss, 
\begin{align}
    \mathcal{L}_{SSL}^{jigsaw} &= \frac{1}{n} \sum_{i=1}^{n}{\mathcal{L}_{CE}(p_i, h_i)}
\end{align}
where this cross-entropy loss is computed between the true permutation-index $p_i$ and the predicted index $h_i$, and averaged over the total number of samples present in the test set. \\

\textbf{Barlow Twins Loss: }
We follow a similar formulation of this loss, as proposed in~\cite{zbontar2021barlow}.
For each image in the test set, $X_i$, we create two differently augmented version of the same as $X^{(1)}_i$ and $X^{(2)}_i$.
Augmented versions are created through various image operations, such as Gaussian blur, grayscale transformation, solarization, etc.
A cross-correlation matrix $\mathcal{C} \in \mathbb{R}^{d\times d}$ is computed between the feature representations of these augmented versions as $\mathbf{x}^{(1)}_i \in \mathbb{R}^{d}$ and $\mathbf{x}^{(2)}_i \in \mathbb{R}^d$ and the following loss function is minimized,
\begin{align}
    \mathcal{L}_{SSL}^{BT} &= \frac{1}{n}\sum_{i=1}^{n}{[\sum_{a=1}^{d}{(1-\mathcal{C}_{aa})^2 + \lambda \sum_{a=1}^{d}{\sum_{b \neq a}{\mathcal{C}_{ab}^2}}}]}
\end{align}
where,
\begin{align}
    \mathcal{C}_{ab} &= \frac{\sum_{i=1}^{n}{\mathbf{x}^{(1)}_{i,a}\mathbf{x}^{(2)}_{i,b}}}{\sqrt{\sum_{i=1}^{n}{(\mathbf{x}^{(1)}_{i,a}})^2} \sqrt{\sum_{i=1}^{n}{(\mathbf{x}^{(2)}_{i,b}})^2}}
\end{align}
Here, the main idea is to make the diagonal term of the cross-correlation matrix 1, so that the embedding becomes invariant to any distortion.
Additionally, the off-diagonal terms are pushed towards zero to de-correlate the different vector components of the embedding~\cite{zbontar2021barlow}.
It has been reported to be particularly successful in image classification problems in the low-data regime.

In our implementation, these self-supervised losses only become effective to adapt the pre-trained base model on the test sets during retrieval.
Thus, we introduce an additional auxiliary classifier with existing pre-trained base-model~(on top of the semantic embedding~(300-d) of SnMpNet) to compute the preferred loss-variant.
However, this additional classifier is required only for RotNet and Jigsaw loss computations; for Barlow Twins, we leverage the 300-d embeddings directly from SnMpNet to compute the $\mathcal{C}$-matrix.
%We choose to introduce this classifier on top of the semantic embedding~(300-d) of SnMpNet.}
%Hypothetically, this auxiliary classifier can be introduced directly after the feature extractor~(SE-ResNet-50, in this case).
%\color{red}However, we choose to introduce this after the semantic projection layer of SnMpNet. \color{black}
Next, we discuss our adaptation or test-time training in detail.

%In our formulation, we modify the SnMpNet training and append one such SSL-loss component to make the network more generalizable towards unknown or unstructured data.
%While training, the model is optimized by minimizing the combination of all three losses, and is updated end-to-end.
%The combined loss becomes,
%\begin{align}
%    \mathcal{L}_{v-SnMpNet} &= \mathcal{L}_{Mp} + \mathcal{L}_{Sn} + \alpha \mathcal{L}_{SSL}^{v}, \nonumber \\
%    & \text{ where }v \in \{rotnet, jigsaw, BT\}
%\end{align}
%Here $\alpha$ is the trade-off factor which balances the loss functions of base SnMpNet and the additional SSL-loss component.

\subsection{Test-time Training~(TTT) for UCDR}
Here, we discuss the proposed test-time training or adaptation framework to address the UCDR in data-efficient manner. 
In our setup, since the training data does not have samples from many numbers of domains or categories, the generalization ability of the base model SnMpNet would be low.
In other words, the network may be biased towards the job of sketch-based image retrieval, in case sketch and image are the domains present during training. 
But its performance degrades~(details in the experiments section, Table~\ref{tab_ucdr_limited_data}) when a cartoon or painting is presented as query.
We hypothesize that under such a condition, any information or clue extracted from test query may help the network to adapt and retrieve better for that query domain.

Towards that goal, we propose to perform a single-step parameter update of the network using the gradient computed through any of the SSL-loss components discussed before.
Thus, for any test sample $X_{te}$, we perform a forward-propagation to compute the $\mathcal{L}_{SSL}^{v}, v \in \{rotnet, jigsaw, BT\}$, and then compute the corresponding gradients and back-propagate through the network components~(SE-ResNet-50 based feature extractor, linear projection layer for learning the semantic-embedding and the auxiliary classifier) just once to tune the  already trained SnMpNet on the basis of the test sample itself.
We make the final inference on the retrieval list for the query sample, using this updated model instead of the previously-trained SnMpNet.
The advantage of this proposed test-time adaptation is that any pre-trained retrieval network can be used as the base model in this case, and no modification in the training strategy or additional data / computation during training is required.
%Such adaptation towards the unknown test samples provides a better generalization in the low-data regime, which we can observe from the experimental data presented in Section~\ref{sec_results}.
\subsubsection{Proposed SnMpNet-variants}
Depending on the SSL-loss component used for test-time adaptation of the model, we propose three variants of SnMpNet as described below:
\begin{enumerate}
    \item rotation-SnMpNet: During test-time, the test samples are rotated as stated previously, to create an augmented set.
    %\color{red}A linear projection layer on top of the 2048-dimensional feature embedding of \emph{SnMpNet} predicts probabilities of the 4 rotation angles. \color{black}
    The 300-d semantic embeddings of the samples in this set are extracted from pre-trained SnMpNet and are fed to the auxiliary classifier to perform the task of classifying their corresponding rotation angles.
    The network parameters are updated only once on the basis of $\mathcal{L}_{SSL}^{rotnet}$ computed over this newly generated test set.
    \item jigsaw-SnMpNet: Here, during test-time training, only $\mathcal{L}_{SSL}^{jigsaw}$ loss will be computed with pre-trained SnMpNet + auxiliary classifier.
    The test samples are resized, and jigsaw images are created for the same. 
    %Similar to rotation-SnMpnet, a \color{red}linear projection layer predicts the permutation index of the jigsaw image. \color{black}
    \item BT-SnMpNet: Finally, for this variant, noisy or distorted test samples are generated to compute the cross-correlation matrix $\mathcal{C}$, using 
    their 300-d semantic embeddings from pre-trained SnMpNet. %$\mathcal{C}$ between the 2 augmentations is computed using the \color{red}300-dimensional semantic embedding output of \emph{SnMpNet}. \color{black}
    We compute $\mathcal{L}_{SSL}^{BT}$ based on $\mathcal{C}$ and accordingly update the model.
\end{enumerate}
Thus, the proposed test-time training methodology is very easy and can be seamlessly used with any trained cross-domain retrieval model without any modification in its architecture or training process.
Now, with this discussion, we'll move on to the experimental validation in the next section.
\color{black}Our overall approach with rotation-SnMpNet is illustrated in Figure~\ref{fig_network} for reference. \color{black}

\section{Experiments}
\label{sec_results}
\begin{table*}[ht!]
\footnotesize
\begin{center}
\begin{tabular}{|c|c|c|c|c|c|c|}
\hline
\multirow{2}{*}{Query Domain} & \multirow{2}{*}{Training Domains} & \multirow{2}{*}{Method} & \multicolumn{2}{c|}{\emph{Unseen}-class Search Set} & \multicolumn{2}{c|}{\emph{Seen}+\emph{Unseen}-class Search Set} \\
\cline{4-7}
& & & mAP@200 & Prec@200 & mAP@200 & Prec@200\\
\hline
\multirow{5}{*}{\emph{Painting}} & \emph{Real}, \emph{Sketch}, \emph{Infograph}, & \multirow{2}{*}{SnMpNet} & \multirow{2}{*}{0.4031} & \multirow{2}{*}{0.3332} & \multirow{2}{*}{0.3635} & \multirow{2}{*}{0.3019} \\
& \emph{QuickDraw}, \emph{Clipart} & & & & &\\
\cline{2-7}
& \multirow{4}{*}{\emph{Sketch}, \emph{Image}} & \emph{SnMpNet} & 0.3827 & 0.3167 & 0.3480 & 0.2842\\
& & \emph{rotation-SnMpNet} & 0.3880 & 0.3219 & \textbf{0.3508} & 0.2892\\
& & \emph{jigsaw-SnMpNet} & 0.3807 & 0.3154 & 0.3441 & 0.2829\\
& & \emph{BT-SnMpNet} & \textbf{0.3932} & \textbf{0.3337} & 0.3481 & \textbf{0.2905}\\
\hline
\multirow{5}{*}{\emph{Clipart}} & \emph{Real}, \emph{Sketch}, \emph{Infograph}, & \multirow{2}{*}{\emph{SnMpNet}} & \multirow{2}{*}{0.4198} & \multirow{2}{*}{0.3323} & \multirow{2}{*}{0.3765} & \multirow{2}{*}{0.2959} \\
& \emph{QuickDraw}, \emph{Painting} & & & & &\\
\cline{2-7}
& \multirow{4}{*}{\emph{Sketch}, \emph{Image}} & \emph{SnMpNet} & 0.3318 & 0.2539 & 0.2923 & 0.2172 \\
& & \emph{rotation-SnMpNet} & \textbf{0.3586} & \textbf{0.2811} & \textbf{0.3173} & \textbf{0.2432}\\
& & \emph{jigsaw-SnMpNet} & 0.3579 & 0.2803 & 0.3158 & 0.2425 \\
& & \emph{BT-SnMpNet} & 0.3465 & 0.2744 & 0.2987 & 0.2271 \\
\hline
\multirow{5}{*}{\emph{QuickDraw}} & \emph{Real}, \emph{Sketch}, \emph{Infograph}, & \multirow{2}{*}{\emph{SnMpNet}} & \multirow{2}{*}{0.1736} & \multirow{2}{*}{0.1284} & \multirow{2}{*}{0.1512} & \multirow{2}{*}{0.1111} \\
& \emph{Clipart}, \emph{Painting} & & & & &\\
\cline{2-7}
& \multirow{4}{*}{\emph{Sketch}, \emph{Image}} & \emph{SnMpNet} & 0.1845 & 0.1471 & 0.1551 & 0.1241 \\
& & \emph{rotation-SnMpNet} & \textbf{0.1931} & \textbf{0.1511} & \textbf{0.1600} & \textbf{0.1246}\\
& & \emph{jigsaw-SnMpNet} & 0.1905 & 0.1498 & 0.1577 & 0.1227  \\
& & \emph{BT-SnMpNet} & 0.1515 & 0.1231 & 0.1082 & 0.0819 \\
\hline
\multirow{5}{*}{\emph{Infograph}} & \emph{Real}, \emph{Sketch}, \emph{Clipart}, & \multirow{2}{*}{\emph{SnMpNet}} & \multirow{2}{*}{0.2079} & \multirow{2}{*}{0.1717} & \multirow{2}{*}{0.1800} & \multirow{2}{*}{0.1496} \\
& \emph{QuickDraw}, \emph{Painting} & & & & &\\
\cline{2-7}
& \multirow{4}{*}{\emph{Sketch}, \emph{Image}} & \emph{SnMpNet} & 0.1660 & 0.1322 & 0.1358 & 0.1071 \\
& & \emph{rotation-SnMpNet} & \textbf{0.1952} & \textbf{0.1592} & \textbf{0.1638} & \textbf{0.1331}\\
& & \emph{jigsaw-SnMpNet} & 0.1904 & 0.1539 & 0.1592 & 0.1281 \\
& & \emph{BT-SnMpNet} & 0.1634 & 0.1326 & 0.1322 & 0.1034 \\
\hline
\multirow{5}{*}{\textbf{\emph{Average}}} & 5/6 DomainNet domains & \emph{SnMpNet} & 0.3011 & 0.2414 & 0.2678 & 0.2146\\
\cline{2-7}
& \multirow{4}{*}{\emph{Sketch}, \emph{Image}} & \emph{SnMpNet} & 0.2663 & 0.2125 & 0.2328 & 0.1832\\
& & \emph{rotation-SnMpNet} & \textbf{0.2837} & \textbf{0.2283} & \textbf{0.2480} & \textbf{0.1975}\\
& & \emph{jigsaw-SnMpNet} & 0.2799 & 0.2248 & 0.2442 & 0.1940\\
& & \emph{BT-SnMpNet} & 0.2636 & 0.2160 & 0.2218 & 0.1757\\
\hline
\end{tabular}
\end{center}
\caption{Performance of SnMpNet and the proposed test-time adaptation based variants for data-efficient UCDR.}
\label{tab_ucdr_limited_data}
\end{table*}
In this section, we discuss the experimental validation of proposed test-time training strategy to address data-efficient UCDR. 
As we mentioned before, SnMpNet~\cite{paul2021universal} is our base model throughout all the experiments and we also analyse the performance of each of the variants discussed in the previous section.
Specifically, we report our results for two different test-cases:
(1) Data-efficient UCDR: where the training set for the SnMpNet is small~(only two training domains), as presented in~\cite{dey2019doodle}\cite{dutta2020adaptive} etc. instead of~\cite{paul2021universal}; and
(2) Traditional UCDR: where the training set contains large-scale multi-domain data, as in~\cite{paul2021universal}.
%(3) Cross-dataset UCDR: where the training data is large~\cite{paul2021universal}, but the test domains belong to a different dataset, thus bringing larger domain-gap in the picture.
First, we briefly introduce the datasets used for this analysis.

%Here, we present the experimental evaluation of our proposed test-time training strategies on top of the pretrained \emph{SnMpNet}. To the best of our knowledge, this is the first such work in this direction, so we only have \emph{SnMpNet's} performance for comparison as a baseline. In Section~\ref{sec_cross_dataset}, we extend our evaluation to the even more challenging cross-dataset generalization setting. We begin with a brief description of the datasets used for our work. 

\subsection{Datasets}
We experimented with two large-scale datasets: \\
\textbf{Sketchy extended~\cite{Sketchy}} contains $75,471$ sketches and $73,002$ images across $125$ categories. 
%We only use it for cross-dataset testing in~\ref{sec_cross_dataset}. 
Out of these total $125$-categories, $93$, $11$, and $21$-classes are used for training, validation, and testing, respectively~(following the ZS-SBIR setup~\cite{dey2019doodle}).
The presence of annotated samples from only two domains makes this dataset an excellent choice for data-efficient generalized retrieval scenario.
Thus, we largely use this dataset for pre-training the base model, and unknwon samples from other domains~(except sketch and image from DomainNet) to validate the effectiveness of proposed test-time adaptation in data-efficient manner.
%Since this contains data from only two domains, thus it provides a real-time training scenario, where collecting multi-domain data as part of the training is challenging.
%While evaluating our approach for Low-data UCDR, we utilize samples~(which belong to a domain other than sketch and image, as well as from an unknown category) from another dataset DomainNet(details in the next paragraph).
%This is to ensure a significant domain-wise and category-wise difference between the query, and the training data - which is the true objective of a generalized retrieval.
% ********************************************

\textbf{DomainNet~\cite{Domainnet}} has approximately $6,00,000$ samples from $345$ categories, collected in six domains, namely, \emph{Clip-art}, \emph{Sketch}, \emph{Real}, \emph{Quickdraw}, \emph{Infograph}, and \emph{Painting}.
Following~\cite{Cumix}, $245$ and $55$ classes are used for training and validation~\cite{Cumix}, respectively.
To simulate the unknown domain, we leave one domain out and rests are used for training.
During testing, the samples from this left-out domain is used as query.
We train the base model using this setting for evaluating the traditional UCDR only.
%It is to be noted that for UCDR and Cross-dataset UCDR, we use training samples from DomainNet, but for Cross-dataset UCDR, the test samples belong to Sketchy-dataset. 

%Since we are experimenting UCDR under a low-data regime, thus we propose a novel cross-dataset train-test split for the same.
%The details of the same has been elaborated in the beginning of each experiment for ease of understanding.
It is not be noted that, for fair comparison, we have always reported results of SnMpNet and proposed variants on the same training and test data-splits.

%the test set is formed with $45$ \emph{unseen} classes.
%Rest  For more details on the UCDR protocol with DomainNet, please refer to~\cite{paul2021universal}. 

\subsection{Implementation details}
\label{subsec_impl}
We use PyTorch 1.1.0 (following~\cite{paul2021universal}) and a single Nvidia Tesla v100 GPU for all experiments. 
Maintaining that any pre-trained baseline model can be used for the purpose, we use the same training parameters for SnMpNet with either just two training domains~(for data-efficient UCDR), or with five domains~(standard UCDR).
%All three SnMpNet-variants are optimized with the training and validation-split of the respective datasets, as discussed above.
For our test-time adaptation, we update the pre-trained model for single-iteration per query sample.
To prevent huge parameter divergence from the learned distribution, we use a lower learning rate (1e-6 to 1e-5) and use SGD optimizer with weight decay and Nesterov momentum of 0.9, with a batch-size of $32$.
This batch is created by applying standard data augmentation techniques, like random resized-crop, horizontal-flip, and color-jitter on the query sample, during the single iteration of training at test-time.
%A batch size of 64 is maintained for \emph{Standard-TTT} and $32$ for \emph{Online-TTT} (see Section~\ref{subsec:std_v_online}).
%During inference, we just update the trained model for a single iteration per sample (as we find this is usually sufficient for parameter adaptation) based on the available test sets. 
%From our experiments, we don't observe any major improvement by training for more iterations. 
%To prevent parameter divergence from the learned distribution on the training samples, we use a lower learning rate (1e-6 to 1e-5) than what was used originally for training. 
%Also, we employ different learning rates for the final projection layers compared to the rest of the network. Particularly, the backbone is learned at $\nicefrac{1}{10}$ of the learning rate for the projection layer. 

\subsection{Test-time Training for Data-efficient UCDR}
First, we analyze the effect of limited training data for UCDR.
Towards that goal, we pre-train the SnMpNet model with the training-split of Sketchy-extended dataset~\cite{dey2019doodle}, and test on the additional domain query data from DomainNet~\cite{paul2021universal}.
This model is treated as the base-model in data-efficient UCDR scenario for the rest of the paper.
We compare this performance with the results reported in~\cite{paul2021universal}, where the same model is trained with multi-domain data from DomainNet. 
Following~\cite{paul2021universal}, these results are reported in the form of mAP@200 and Prec@200 in Table~\ref{tab_ucdr_limited_data}.
Also, we perform experiments for two different configurations of the search set: (1) when the search set contains samples from unseen categories only; and (2) when both seen and unseen categories samples are present in the search set. We exclude the \emph{Sketch} domain from DomainNet as query in this experiment, as it is already present in the training set of Sketchy-extended and hence isn't effectively unseen-domain to the model.

As we can observe from Table~\ref{tab_ucdr_limited_data}, the performance of SnMpNet drops significantly when the number of training domains becomes restricted.
This result is on par with our expectation, since traditional SnMpNet is not originally designed to learn generalization in a data-efficient manner.
This regression in performance has been observed for all unseen query domains, except for \emph{Quickdraw}.
This may be because the contents of \emph{Quickdraw} is basically \emph{Sketch}~(but very abstract in nature); thus, SnMpNet can retrieve for this case easily, compared to other drastically different~(eg. \emph{Painting} or \emph{Infograph}) query domains.
Thus, this observation justifies the need for further effort in this direction.
% ************************
% ***********************

%\textbf{Effectiveness of TTT for low-data UCDR: }
Next, we explore the effectiveness of test-time training in data-efficient learning context.
We observe the performance of all 3 \emph{SnMpNet} variants under limited training domains in Table~\ref{tab_ucdr_limited_data}.
We can observe that for all 4 query domains (sketch excluded), test-time training has improved the performance of SnMpNet~\cite{paul2021universal}.
Particularly, \emph{rotation-SnMpnet} significantly outperforms the base model on all four query domains and almost matches the original \emph{SnMpNet}~\cite{paul2021universal} performance~(trained on all five-domains).
These results are encouraging and support our previous hypothesis that any information/hint extracted from the unknown test sample could help improve the model's generalization performance under such a challenging condition.
Also, the proposed adaptation does not require any modification on the pre-trained base network, which makes it easy to integrate with any data-efficient base model.

\subsection{Test-time Training for Traditional UCDR}
\label{subsec_ucdr_results}
Next, we study the effectiveness of such test-time adaptation for traditional UCDR.
Following~\cite{paul2021universal}, we assume that sufficient multi-domain data is available to the model for training, and aim to explore if proposed test-time adaptation can further improve the performance of the model for UCDR.
Thus, the \emph{SnMpNet} is pre-trained with training samples from five domains~(leaving one out) from DomainNet, and used as base model.
While testing the query from unknown domain, we apply the test-time adaptation discussed in Section~\ref{sec_proposed}.
%Here, the pre-trained \emph{SnMpNet} uses training data from DomainNet.
%o simulate the ``unknown domain and unknown class" query-data notion, one domain is left out during training, and we use samples from unused category of this held-out domain for testing.
We summarize our observations in Table~\ref{tab_ttt_ucdr}.
\begin{table*}[ht!]
\footnotesize
\begin{center}
% \begin{tabular}{| >{\raggedright\arraybackslash}p{0.05\textwidth}| >{\centering\arraybackslash}p{0.07\textwidth}| >{\centering\arraybackslash}p{0.055\textwidth}| >{\centering\arraybackslash}p{0.05\textwidth}| >{\centering\arraybackslash}p{0.055\textwidth}| >{\centering\arraybackslash}p{0.05\textwidth}|}
\begin{tabular}{|c|c|c|c|c|c|}
\hline
Query Domain & \multirow{2}{*}{Method} & \multicolumn{2}{c|}{\emph{Unseen-class Search Set}} & \multicolumn{2}{c|}{\emph{Seen+Unseen-class Search Set}} \\
\cline{3-6}
 & & mAP@200 & Prec@200 & mAP@200 & Prec@200\\
\hline
\multirow{4}{*}{\emph{Sketch}} & \emph{SnMpNet} & 0.3007 & 0.2432 & 0.2624 & 0.2134 \\
& \emph{rotation-SnMpNet} & 0.2959 & 0.2394 & 0.2624 & 0.2129 \\
& \emph{jigsaw-SnMpNet} & 0.2929 & 0.2371 & 0.2632 & 0.2134 \\
& \emph{BT-SnMpNet} & 0.2993 & \textbf{0.2479} & 0.2440 & 0.1987 \\
\cline{1-6}
\multirow{4}{*}{\emph{Quickdraw}} & \emph{SnMpNet} & 0.1736 & 0.1284 & 0.1512 & 0.1111 \\
& \emph{rotation-SnMpNet} & 0.1683 & 0.1244 & 0.1494 & 0.1094\\
& \emph{jigsaw-SnMpNet} & 0.1689 & 0.1250 & 0.1498 & 0.1100 \\
& \emph{BT-SnMpNet} & 0.1613 & \textbf{0.1298} & 0.1134 & 0.0785 \\
\cline{1-6}
\multirow{4}{*}{\emph{Painting}} & \emph{SnMpNet} & 0.4031 & 0.3332 & 0.3635 & 0.3019 \\
& \emph{rotation-SnMpNet} & 0.3997 & 0.3301 & \textbf{0.3707} & \textbf{0.3064}\\
& \emph{jigsaw-SnMpNet} & 0.3992 & 0.3301 & 0.3675 & 0.3041 \\
& \emph{BT-SnMpNet} & \textbf{0.4072} & \textbf{0.3494} & 0.3615 & 0.3042 \\
\cline{1-6}
\multirow{4}{*}{\emph{Infograph}} & \emph{SnMpNet} & 0.2079 & 0.1717 & 0.1800 & 0.1496 \\
& \emph{rotation-SnMpNet} & 0.2058 & 0.1695 & 0.1815 & 0.1508 \\
& \emph{jigsaw-SnMpNet} & 0.2053 & 0.1692 & \textbf{0.1819} & \textbf{0.1511} \\
& \emph{BT-SnMpNet} & 0.1903 & 0.1597 & 0.1502 & 0.1229 \\
\cline{1-6}
\multirow{4}{*}{\emph{Clipart}} & \emph{SnMpNet} & 0.4198 & 0.3323 & 0.3765 & 0.2959 \\
& \emph{rotation-SnMpNet} & 0.4171 & 0.3295 & 0.3786 & 0.2978\\
& \emph{jigsaw-SnMpNet} & 0.4167 & 0.3298 & \textbf{0.3835} & \textbf{0.3020} \\
& \emph{BT-SnMpNet} & \textbf{0.4281} & \textbf{0.3472} & 0.3790  & 0.2962 \\
\cline{1-6}
\multirow{4}{*}{\emph{{\emph{Average}}}} & \emph{SnMpNet} & 0.3010 & 0.2418 & 0.2667 & 0.2144 \\
& \emph{rotation-SnMpNet} & 0.2974 & 0.2386 & 0.2685 & 0.2155 \\
& \emph{jigsaw-SnMpNet} & 0.2966 & 0.2382 & \textbf{0.2692} & \textbf{0.2161} \\
& \emph{BT-SnMpNet} & 0.2972 & \textbf{0.2468} & 0.2496 & 0.2001 \\
\hline
\end{tabular}
\end{center}
\vspace{-1em}
\caption{Comparison of \emph{SnMpNet} and its \emph{TTT}-variants for UCDR protocol on DomainNet. }
\label{tab_ttt_ucdr}
\end{table*}
We see that at least one of the proposed variants outperforms \emph{SnMpNet} on \emph{Painting}, and \emph{Clipart} domains for both configurations of the search set. 
However, performance drops for \emph{Sketch}, \emph{Infograph} and \emph{Quickdraw} by small amounts. 
This also impacts the overall \emph{Average} performance. 
Only BT-SnMpNet outperforms \emph{SnMpNet} on Prec@200 for unseen-class search set.
We note this to be a limitation of test-time adaptation in traditional UCDR.
In case sufficient training data is available during training, the proposed self-supervision-based test-time adaptation cannot improveme on the pre-trained base model's retrieval performance.
It can be noted that such a failure case is on the similar line of analysis for TTT reported by TTT++~\cite{ttt++} in the context of traditional image-classification.

With these results, we now move on to the in-depth analysis of the proposed approach in the next section.

\section{Analysis}
\label{sec_analysis}
Here, we provide reasoning behind the design choices for the proposed test-time adaptation, and ablate different learning components to understand the contribution of each.
We choose the \emph{rotation-SnMpNet} variant for this analysis, since it has demonstrated improvement consistently~(refer to Table~\ref{tab_ucdr_limited_data}).
The results in this section are reported with data-efficient pre-trained base-model, where the number of training domains is two for pre-training.

\subsection{Ablation Studies}
\label{subsec:ablation}
We first analyze the impact of different design parameters during test-time training of \emph{SnMpNet}. 
We primarily identify the learning rate, embedding-dimension, optimizer configuration, no. of iterations during test-time update as some of the crucial parameters effecting the final performance of model.
We summarize the effect of each of these parameters in Table~\ref{tab_ablation}, using \emph{Infograph} as the query domain.
We observe that the test-time adaptation strategy is almost robust to the number of update-iterations, optimizer configuration or embedding-dimension etc., since the reported mAP@200 and Prec@200 remain similar against these conditions.
However, learning rate of this adaptation seems to affect the retrieval results noticably. 
We see that a learning rate of 1e-6 for the base model and 1e-5 for the auxiliary classifier performs much better than 1e-4 and 1e-3, respectively.
For context, pre-trained SnMpNet final learning rate is 1e-6 for our experiments.
This finding is on par with recommendations in~\cite{sun2020test} of TTT learning rate being at the same scale of the pre-trained model's final epoch of training.
%We choose \emph{rotation-SnMpNet} since it has the best performance across the 3 variants in Table~\ref{tab_ucdr_limited_data} and evaluate it for UCDR on \emph{Infograph} domain. We study the impact of number of iterations on a test sample, optimizer configuration, learning rate, and embedding dimension on the performance of \emph{rotation-SnMpNet}. As can be seen in Table~\ref{tab_ablation}, the training paradigm is fairly robust to all these hyperparameters except for learning rate. This is expected as \cite{sun2020test} recommends the initial learning rate for TTT to be set to that at the last epoch during training. This parameter is 1e-6 for the last epoch of \emph{SnMpNet} training. Hence a learning rate of 1e-6 for the backbone and 1e-5 for the auxiliary classifier performs much better than 1e-4 and 1e-3, respectively. The auxiliary classifier can either take the 2048-dimensional ResNet feature embedding or the 300-dimensional semantic embedding as input for predicting the rotation angles. This has very minimal impact on the performance. Contrary to the findings of \cite{sun2020test} in the classification setting, here, the SGD optimizer works equally with and without Nesterov momentum (0.9) and L2 decay (5e-4). 

\begin{table*}[ht!]
\footnotesize
\begin{center}
\begin{tabular}{|c|c|c|c|c|c|c|c|c|c|}
\hline
Query Domain & \multirow{2}{*}{Iter.} & \multirow{2}{*}{Mom.} & \multirow{2}{*}{lrc/lrb} & \multirow{2}{*}{Emb. Dim} & \multirow{2}{*}{L2 decay} & \multicolumn{2}{c|}{\emph{Unseen-class Search Set}} & \multicolumn{2}{c|}{\emph{Seen+Unseen-class Search Set}} \\
\cline{7-10}
 & & & & & &mAP@200 & Prec@200 & mAP@200 & Prec@200\\
\hline
\multirow{5}{*}{\emph{Infograph}} & 1 & \checkmark & 1e-3/1e-4 & 300 & \checkmark & 0.1869 & 0.1505 & 0.1561 & 0.1250 \\
\cline{2-10}
& 1 & \xmark & 1e-5/1e-6 & 300 & \xmark & 0.1952 & 0.1591 & 0.1638 & 0.1330 \\
\cline{2-10}
& 3 & \checkmark & 1e-5/1e-6 & 300 & \checkmark & 0.1940 & 0.1583 & 0.1626 & 0.1321 \\
\cline{2-10}
& 1 & \checkmark & 1e-5/1e-6 & 2048 & \checkmark & 0.1950 & 0.1587 & 0.1634 & 0.1325 \\
\cline{2-10}
& 1 & \checkmark & 1e-5/1e-6 & 300 & \checkmark & \textbf{0.1952} & \textbf{0.1592} & \textbf{0.1638} & \textbf{0.1331} \\
\hline
\end{tabular}
\end{center}
\vspace{-1em}
\caption{Ablation Study for proposed test-time adaptation technique with \emph{rotation-SnMpNet} for data-efficient UCDR evaluation. \emph{Infograph} from DomainNet dataset is used as query domain. [Column description: Iter: number of iterations during test-time adaptation; Mom: Nesterov momentum; lrc/lrb: learning rate used for auxilliary classifier / pre-trained base model during adaptation; Emb. Dim: embedding dimension fed to the axilliary classifier; L2-Decay: is used in optimizer or not.]}
\label{tab_ablation}
\end{table*}

\subsection{Standard v Online TTT}
\label{subsec:std_v_online}
Here, we explore variants of the test-time update strategy based on the literature~\cite{sun2020test}.
We refer to the adaptation followed in this work as \emph{Standard} adaptation, where the base-model parameters are updated based on each query sample, and then discarded for the next query.
Essentially, for each query, the original pre-trained SnMpNet is being adapted and accordingly retrieval is performed.
We also explore the retrieval results using the \emph{online}-version~\cite{sun2020test} of adaptation, where the updates from each sample get accumulated and the pre-trained base model keeps updating itself sequentially for each incoming query.
This strategy has been reported to be effective in the classification setting for gradually changing distribution shifts at test time.
%It has been reported to be an effective process for classification.
However, the evaluation summarized in Table~\ref{tab_std_online} clearly depict that it is not as effective as the standard adaptation process for data-efficient UCDR.
This may be because that UCDR is much more complex compared to traditional classification problem.
In UCDR, each query can potentially belong to a different domain and category, and thus a cumulative update from each of such samples can lead to complete parameter divergence for the model.
Thus the assumption of gradually changing distribution shifts that hold for \emph{traditional} TTT benchmarks~\cite{sun2020test}, won't be valid for UCDR. 
%where samples at test time come from previously unseen classes and domains.
%In the rest of this paper, we have reported results using the \emph{Standard} version of TTT, where the model parameters $\theta$ obtained after making an update on $x_t$ are discarded after making the prediction for $x_t$. Here, similar to \cite{sun2020test}, we additionally report results using the \emph{Online} version of TTT, which has been shown to be effective in the classification setting for gradually changing distribution shifts at test time. In \emph{TTT-Online}, the test samples arrive online sequentially, and the parameters $\theta$ for $x_t$ are initialized with $\theta(x_{t-1})$. For analysis, we pick \emph{rotation-SnMpNet} and report its performance for the \emph{UCDR} protocol on DomainNet in Table~\ref{tab_std_online}. As we can observe, this paradigm leads to a complete parameter divergence, and the model performs much worse than the \emph{TTT-Standard} and the original \emph{SnMpNet}. We hypothesize that this is because the assumption of gradually changing distribution shifts that hold for \emph{traditional} TTT benchmarks, won't be valid in the UCDR case where samples at test time come from previously unseen classes and domains. Similar to \cite{sun2020test}, we also shuffle the test set before \emph{TTT-Online} to prevent \emph{ordering artifacts}.

\begin{table*}[ht!]
\footnotesize
\begin{center}
\begin{tabular}{|c|c|c|c|c|c|}
\hline
Query Domain & \multirow{2}{*}{TTT-version} & \multicolumn{2}{c|}{\emph{Unseen-class Search Set}} & \multicolumn{2}{c|}{\emph{Seen+Unseen-class Search Set}} \\
\cline{3-6}
 & & mAP@200 & Prec@200 & mAP@200 & Prec@200\\
\hline
\multirow{2}{*}{\emph{Sketch}} & \emph{Standard} & 0.2959 & 0.2394 & 0.2624 & 0.2129 \\
& \emph{Online} & 0.2517 & 0.1904 & 0.2223 & 0.1685 \\
\hline
\multirow{2}{*}{\emph{Quickdraw}} & \emph{Standard} & 0.1683 & 0.1244 & 0.1494 & 0.1094 \\
& \emph{Online} & 0.0988 & 0.0733 & 0.0867 & 0.0649 \\
\hline
\multirow{2}{*}{\emph{Painting}} & \emph{Standard} & 0.3997 & 0.3301 & 0.3707 & 0.3064 \\
& \emph{Online} & 0.3405 & 0.2713 & 0.3153 & 0.2506 \\
\hline
\multirow{2}{*}{\emph{Infograph}} & \emph{Standard} & 0.2058 & 0.1695 & 0.1815 & 0.1508 \\
& \emph{Online} & 0.1340 & 0.1006 & 0.1173 & 0.0880 \\
\hline
\multirow{2}{*}{\emph{Clipart}} & \emph{Standard} & 0.4171 & 0.3295 & 0.3786 & 0.2978 \\
& \emph{Online} & 0.3320 & 0.2501 & 0.2947 & 0.2219 \\
\hline
\end{tabular}
\end{center}
\vspace{-2em}
\caption{Comparison of \emph{Standard} and \emph{Online} versions of \emph{rotation-SnMpNet} for UCDR protocol on DomainNet.}
\label{tab_std_online}
\end{table*}

% **************************************************
\section{Conclusion}
\label{sec_conclusion}
%In this work, we explored the scope of 
In this work, we proposed test-time training heuristics for data-efficient UCDR task.
We leveraged self-supervision learning techniques to adapt any pre-trained data-efficient retrieval model to generalize across any unknown domain and/or category during inference.
To the best of our knowledge, though test-time training has been previously explored in the classification and ZS-SBIR settings, this is the first work exploring data-efficient UCDR using test-time adaptation techniques.
We also reported extensive experiments and in-depth analysis to corroborate our hypothesis.
We have demonstrated that our approach is simple, easy to integrate, and effective for such a challenging retrieval paradigm.
%in 3 different self-supervision techniques. 
%To the best of our knowledge, this is the first such work in this direction. 
%In addition, we also study cross-dataset generalization in the UCDR setting and show how test-time training can massively bridge the domain gap across datasets.
%Extensive experiments and comparisons show the effectiveness of our proposed techniques. 

%To build on our work further, we plan to introduce the self-supervised loss components, especially, RotNet-loss, during the training stage of SnMpNet. We specifically choose Rotnet because of its relative simplicity of implementation and also its superior performance in the low-data regime among the 3 self-supervised losses.
%We hypothesize that this would help the network to learn the RotNet loss dynamics better before being exposed to it at test-time. 
%We can introduce this loss into SnMpNet by simply rotating the mixed image $\mathbf{\tilde{x}}$ into one of 4 angles (0°, 90°, 180°, 270°) and applying all 4 losses on this rotated image.

%%%%%%%%% REFERENCES
{\small
\bibliographystyle{ieee_fullname}
\bibliography{PaperForReview}
}

\end{document}